\documentclass[journal]{IEEEtran}
\usepackage[pdftex]{graphicx}
\usepackage{amsmath}

\usepackage{float}
\usepackage{caption}
\usepackage{array}
\usepackage{algorithmic}
\usepackage{multirow}
\usepackage{array}
\usepackage[export]{adjustbox}
\newcolumntype{P}[1]{>{\centering\arraybackslash}p{#1}}
\usepackage{cite}
\usepackage[font=small,labelsep=period]{caption}
\usepackage{lipsum}
\usepackage{bbding}
\usepackage{dingbat}
\usepackage{rotating}
\usepackage{dblfloatfix}
\usepackage{url}
\usepackage[table,xcdraw]{xcolor}
\usepackage{lipsum}
\usepackage{float}
\usepackage{comment}
\usepackage{soul}
\usepackage{url}
\usepackage{hyperref}
\usepackage{lscape}

\usepackage{graphicx}
\usepackage{epstopdf}
\usepackage[T1]{fontenc}

\ifCLASSINFOpdf
\else
\fi
\usepackage{amsmath}
\usepackage[cmintegrals]{newtxmath}
\usepackage{epstopdf}

\begin{document}
%
\title{Explainable Artificial Intelligence (XAI) for \\Internet of Things: A Survey}

\author{İbrahim~Kök, Feyza Yıldırım Okay, Özgecan Muyanlı, and Suat~Özdemir
\thanks{I. Kok is with the Department of Computer Engineering, Pamukkale University, Denizli,
	TR e-mail: ikok@pau.edu.tr}
\thanks{F. Y. Okay is with the Department of Computer Engineering, Gazi University, Ankara, TR e-mail: feyzaokay@gazi.edu.tr}
\thanks{O. Muyanlı and S. Özdemir are with the Department of Computer Engineering, Hacettepe University, Ankara, TR e-mail: ozgecanmuyanli@gmail.com, ozdemir@cs.hacettepe.edu.tr.}}

\markboth{T\MakeLowercase{his work has been submitted to the} IEEE \MakeLowercase{for possible publication.} C\MakeLowercase{opyright may be transferred without notice, after which this version may no longer be accessible.}}%
{Shell \MakeLowercase{}: Bare Demo of IEEEtran.cls for IEEE Communications Society Journals}
%

\maketitle
\begin{abstract}
Black-box nature of Artificial Intelligence (AI) models do not allow users to comprehend and sometimes trust the output created by such model. In AI applications, where not only the results but also the decision paths to the results are critical, such black-box AI models are not sufficient. Explainable Artificial Intelligence (XAI) addresses this problem and defines a set of AI models that are interpretable by the users. Recently, several number of XAI models have been to address the issues surrounding by lack of interpretability and explainability of black-box models in various application areas such as healthcare, military, energy, financial and industrial domains. Although the concept of XAI has gained great deal of attention recently, its integration into the IoT domain has not yet been fully defined. In this paper, we provide an in-depth and systematic review of recent studies using XAI models in the scope of IoT domain. We categorize the studies according to their methodology and applications areas. In addition, we aim to focus on the challenging problems and open issues and give future directions to guide the developers and researchers for prospective future investigations.
\end{abstract}


\IEEEpeerreviewmaketitle

\section{Introduction}
The Internet of Things (IoT) is a prominent technology that connects smart things/objects, allowing them to communicate with each other and provide better services to users \cite{sobin2020survey}. By managing and controlling its underlying technologies, IoT transforms traditional electronic devices such as sensors, actuators, RFID tags, cell phones etc.  into smart objects. Thus, it empowers objects to see, hear, think and perform certain tasks by enabling them to synchronize and share information with each other \cite{shah2016survey}. In order to improve the quality of life, IoT offers numerous potentials and opportunities in different application areas such as smart cities, smart buildings, smart agriculture, healthcare, finance, military \cite{ngu2016iot}.

IoT allows the generation of massive amounts of data that needs to be fine-grained analysis. While this raw data is meaningless, AI systems make it possible to extract meaningful information and provide insightful decisions that affect human lives (in critical fields such as healthcare or autonomous systems) \cite{mukhopadhyay2021artificial}. In recent years, as in other fields, IoT applications have extensively used AI models to overcome the problems caused by the rapidly increasing amount of data and the number of devices \cite{ghosh2018artificial}. In particular, many AI models are widely employed in autonomous network management, device management, service management and analysis of massive IoT data which require smart decision making with high precision and accuracy \cite{kishor2021artificial}. 

As AI technology becomes more integrated into our daily lives, it becomes increasingly important to comprehend how and why decisions are made. However, as machine learning models become more powerful, they often become more complex but less transparent. These powerful models are generally called 'black-box' and they suffer from opaqueness. In other words, they exclude the internal logic from their users \cite{guidotti2018survey}. Therefore, recently, the concept of Explainable Artificial Intelligence (XAI) has started to attract the attention of researchers to cope with the current challenges and to design more explainable/interpretable AI systems. XAI enables to shine a light on the opaqueness of the black-box models to reveal unseen/hidden information such as feature importance, and correlations between features. They will provide more detailed information about how, why and when they make decisions about the inner workings of black-box models and provides transparency to their users. Thus, users are able to evaluate not only the result but also the input factors affecting the result when making a decision. XAI techniques achieves both explainability and high accuracy when it is applied to powerful and complex models. The studies of \cite{guidotti2018survey,samek2017explainable} emphasize the need for explanations of the human-related issues, not just in computer science but also in cognitive science, philosophy, and psychology. According to these studies, reaching the outcome without any interpretation may result in intentional or unintentional discrimination or trust issues. XAI mitigates these problems by providing verification, improvement, learning and compliance of legislation \cite{samek2017explainable}.

In this study, we envision that the integration of XAI approaches into the IoT domain will reveal a serious research potential in terms of the transparency, explainability and interpretability of the AI and ML models in IoT. However, we see that there is not enough research effort in the literature on this subject. To date, there are several survey papers focused on XAI from a general perspective. 
For example, the studies of \cite{gilpin2018explaining,tjoa2020survey,arrieta2020explainable} explain XAI and focus on XAI concepts, terminology, taxonomy and challenges, whereas the study of \cite{rojat2021explainable} examines XAI for time-series data. However, to the best of our knowledge, there is not any existing survey paper investigating or addressing the use of XAI techniques in IoT domain. To fulfill this gap, in this paper, we provide a comprehensive review of current studies on XAI in the IoT domain.
The main contributions of this paper are as follows:
\begin{itemize}
    \item For early researchers in this area, we explain XAI terminology and techniques in a simple and clear manner.
    \item We present a comprehensive review of the current studies addressing XAI methods in IoT domain.
    \item We highlight the emerging challenges and open issues in XAI from an IoT perspective and outline future research directions. 
\end{itemize}

The remainder of the paper is structured as follows. In Section \ref{Sec2_Xai}, we present the taxonomy, terminology,  and methodology of XAI. In Section \ref{Sec3_xai_inIoT}, we explain the needs and benefits of the synergies between XAI and IoT. In Section \ref{Sec4_xai_in_IoT_Domains}, we review current studies addressing XAI by considering IoT application areas. Section \ref{Sec5_OpenC_FutureD} outlines open challenges and future directions. Finally, Section \ref{Sec6_Conclusion} concludes the paper.

\section{Explainable Artificial Intelligence (XAI)}
\label{Sec2_Xai}
\subsection{Terminology and Definitions}

Interpreting and explaining models are not a new problem in computer science and statistics domains. The difficulty of describing the outcome of an expert system is long lasting issue that has been tackled since the 1970s. With the proliferation of AI applications, especially since 2016, interpreting and explaining models started to attract more attention. Practitioners and academics in the literature interchangeably use the terms XAI and Interpretable Machine Learning (IML). Nevertheless, IML is a subset of XAI that focuses on describing the logic of machine learning algorithms \cite{adadi2018peeking}.

It is hard to give a precise definition for the term of explanation. Although some systems are explainable by nature, an explanation helps to make a system more understandable. Especially determining what is a good explanation or not is a debatable issue in the literature \cite{confalonieri2021historical}. In addition, XAI and IML are two similar concepts that are used interchangeably in the literature \cite{adadi2018peeking}, focusing on explainability and interpretability, respectively. However, there are some fundamental differences between these two terms. To clarify the similarities and differences among the similar terms, we provide brief definitions of the terms as follows \cite{arrieta2020explainable,guidotti2018survey,van2020xai,vilone2020explainable}:

\begin{itemize}
    \item \textbf{Explainability:} It is the ability to provide information about the inner workings of the model.
    \item \textbf{Interpretability:} It is the ability to extend a model or its predictions understandable by humans. It is expressed through transparency.
    \item \textbf{Understandability / Intelligibility:} It is the ability to make a human grasp its function about how it works without having to explain its underlying structure or the algorithmic ways.
   
    \item \textbf{Comprehensibility:} It is the ability to express learned information to make it understandable to humans.
    \item \textbf{Transparency:} It is the opposite of being opaque for a black-box model. A system or a model becomes transparent when it is understandable by humans.
    \item\textbf{Faithfulness:} It is the ability of being consistency of choosing the truly relevant features.
    \item\textbf{Informativeness:} It is the ability of a strategy for explainability to offer end-users with meaningful information.
    \item\textbf{Explicitness:} It is the ability of a method to deliver immidiate and clear explanations.
\end{itemize}

\begin{figure}[hbt]
	\centering
	\includegraphics[scale=0.70]{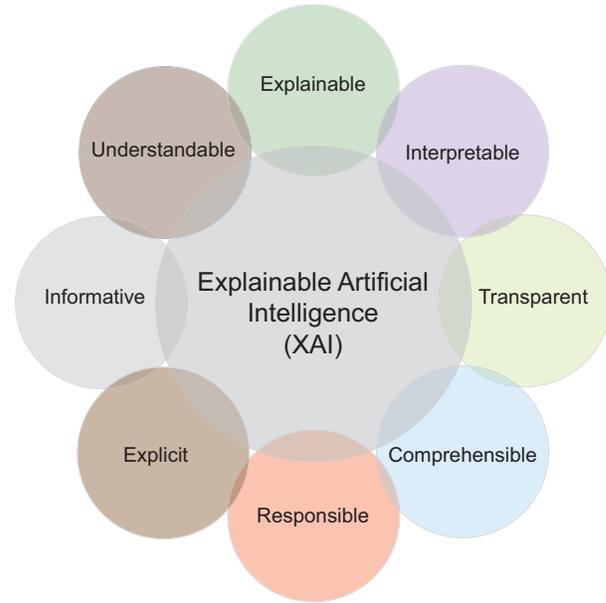}
	\captionsetup{justification=raggedright,singlelinecheck=false}
	\caption{Buzzwords in XAI}
	\label{fig:wordcloud}
\end{figure}

These terms are given in Figure \ref{fig:wordcloud}.
\subsection{XAI Taxonomy}

\begin{figure*}[hbt]
			\centering
			\includegraphics[scale=.50]{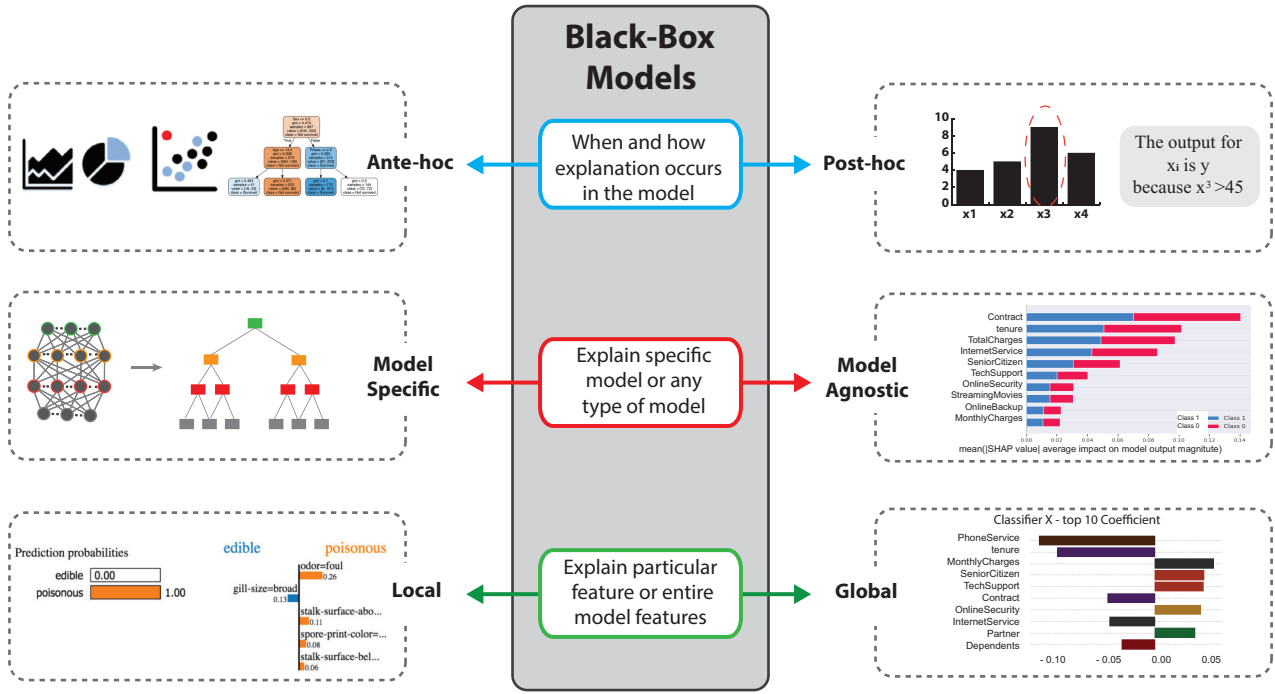}
			\captionsetup{justification=raggedright,singlelinecheck=false}
			\caption{The taxonomy of XAI in different perspectives}
			\label{fig:taxonomy}
		\end{figure*}

The taxonomy of XAI can be classified from a variety of perspectives \cite{carvalho2019machine, arrieta2020explainable}, as shown in Figure \ref{fig:taxonomy}. It is worth point out that there may be overlaps when a method is categorized under this taxonomy. That is, a method can be classified into one or more categories. A method can be classified as post-hoc, model-agnostic, or local, for example. Accordingly, it is more accurate to examine each method separately under its own sub-taxonomy classification. Furthermore, each figure should be considered as an example method in the related group. Different figures can be used to represent the taxonomy.

\subsubsection{Ante-hoc vs. Post-hoc}
An explanation can be provided to a model in pre-training, in-training or post-training phases. Explainable AI can be applied externally in pre-training or post-training phases, or the model itself is intrinsically interpretable during the training which is also called transparent. Ante-hoc methods involve interpreting externally before the training phase or internally during the training phase. At the end of the training phase, the model becomes already explainable. Transparent methods like Decision Tree and Sparse Linear Regression generally give intrinsic explainability about the inner process of the structure. $(ii)$ Post-hoc methods, on the other hand, are applied externally after the training process of decision systems. In addition, post-hoc methods can used to the supplement ante-hoc methods for providing additional information.

\subsubsection{Model-specific vs. Model-agnostic}
XAI can be grouped based on applying to specific model classes, which are model-specific and model-agnostic explanations $(i)$ Model-specific methods are typically designed for one type of model such as Deep Neural Network which is a well-known example of black-box models with a superior prediction performance despite its complex and opaque structure. The disadvantage of model-specific explanations is that there is a limitation in determining a model When the need for the particular type of explanation. $(ii)$ Model-agnostic methods can be applied to any type of model without limiting the model classes. It separates the explanation and model class. Therefore, the explanations become independent of the model type.

\subsubsection{Local vs. Global}
According to the scope of the model, explanations of the decision models can be performed locally or globally. $(i)$ Local explanation describes why and how certain predictions can be generated on a local level. It concentrates on ensuring interpretability by examining single or multiple instances. It is applied especially when predictions are linearly dependent on some features rather than complex dependence on all features. $(ii)$ The goal of global explanation is to characterize the model in its entirety. It seeks a global understanding of which features are more significant and what kinds of relationships are possible between them.

\subsection{XAI Methodologies}
\begin{figure*}[!htb]
			\centering
			\includegraphics[width=7in]{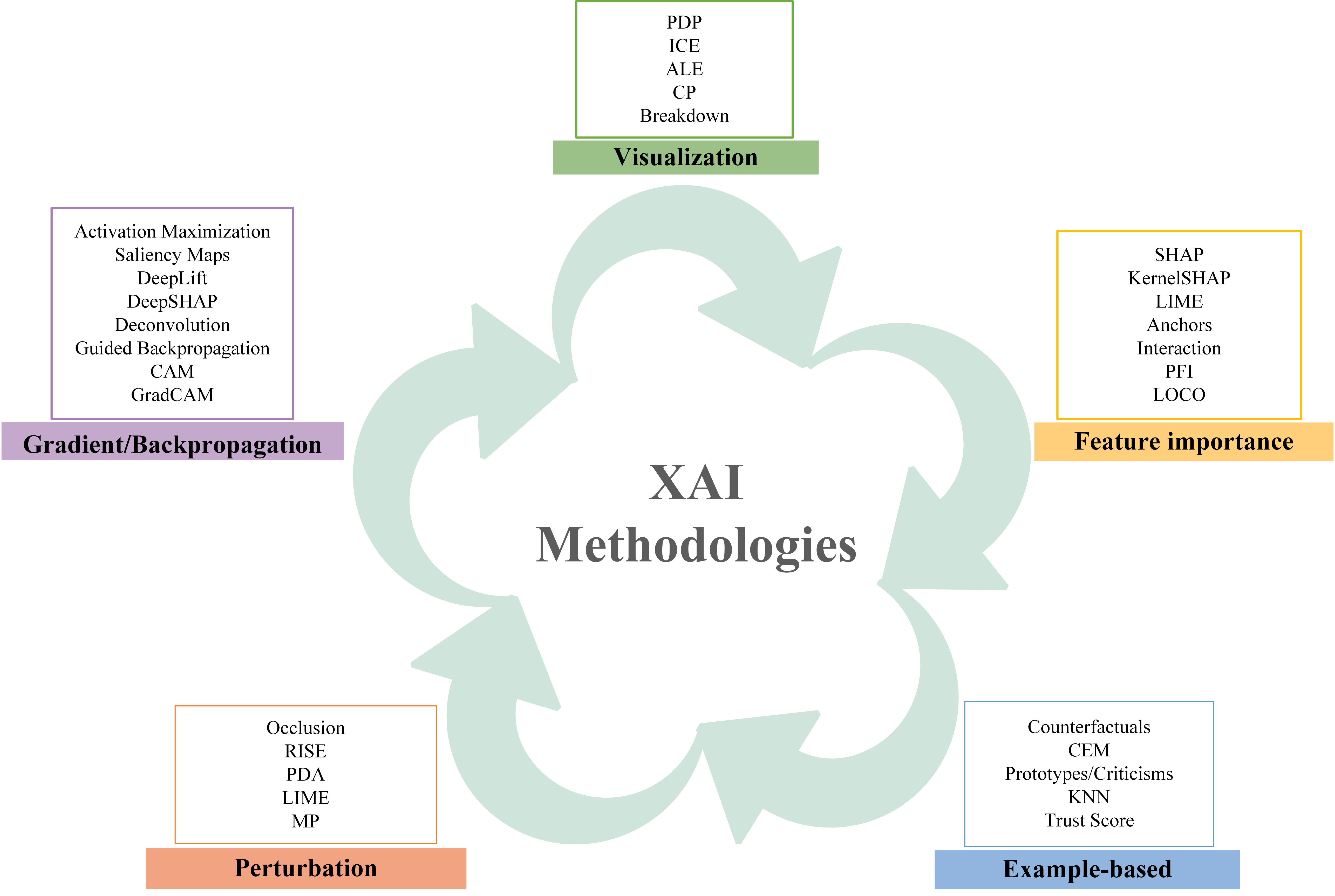}
			\captionsetup{justification=raggedright,singlelinecheck=false}
			\caption{The methodologies of XAI}
			\label{fig:Methodologies}
		\end{figure*}

With the XAI concept gaining popularity in AI, different XAI methods have begun to be developed. In literature, there are different XAI methods to help explain and interpret black-box models. These methods adopt different approaches and provide different interpretations \cite{kamath2021explainable}. Here, we classify them into five different groups as seen in Fig. \ref{fig:Methodologies}.

\subsubsection{Visual Explanation}
Visual explanations cover a set of methods to examine the relationships between input and output or among input attributes, which allows users to understand the contributions of each input to the output. In case the feature set is small, interpreting these correlations is easier for users. Both local and global explanations are supported visually. However, as the feature set becomes large, visual explanations can fail to visualize correlations properly, causing users to misinterpret them. Partial Dependence Plot (PDP) \cite{friedman2001greedy} provides global explanations and reveals the dependency between the input features and output. Individual conditional explanation (ICE) \cite{goldstein2015peeking} plot is a more recent method that's similar to PDP. PDP calculates the mean over the marginal distribution, whereas ICE keeps the entire distribution. Accumulated Local Effects (ALE) \cite{apley2020visualizing} plots take the averages of changes in predictions and accumulate them on local grids. Also, Ceteris Paribus (CP) \cite{kuzba2019pyceterisparibus} plots and Breakdown plots \cite{pontiveros2020explainable} are employed to visualize the influence of features on the model prediction for a specific data instance. 

\subsubsection{Feature-based Explanation}
Feature-based approaches, like visual explanation methods, aim to assess the contribution of features to the model prediction. Furthermore, they consider some factors like type, robustness, and comprehensibility. Explanations can be local or global. Similarly, it can be model-specific or model-agnostic. Shapley values is a game-theoretic approach to determine the feature importance of the model. Then, SHapley Additive exPlanations (SHAP) \cite{lundberg2017unified} method uses Shapley values and it is proposed for local and global explanations. Also, KernelSHAP \cite{lundberg2017unified} is proposed to overcome the conditional expectations issues of SHAP and approximates the calculation of SHAP. Also, local and global surrogate models attempt to explain the prediction of the model. While local surrogate models like Local Interpretable Model-Agnostic Explanations (LIME) \cite{ribeiro2022iot} focus on explaining the individual data instances, global surrogate models focus on the entire model. Anchors \cite{ribeiro2018anchors} have also the ability of local explanations by extracting a set of if-then rules. Feature Interaction \cite{friedman2008predictive} is used for identifying the interaction effect of the pair of feature-output or feature-feature. Permutation Feature Importance (PFI) \cite{galkin2018human} is a global explanation approach based on the idea of shuffling the values of unimportant features does not increase the prediction error. Also, Leave-One-Covariate-Out (LOCO) \cite{lei2018distribution} is another method that involves dropping each variable one at a time, retraining the model, and comparing the following model error to a baseline model that consists of all features.

\subsubsection{Example-based Explanation}
Example-based explanations consist of model-agnostic methods that aim to explain the global or local behavior of a model or its underlying data distribution over certain data instances. Different from visual explanation or feature importance, they assist users to build compact models of the decision model. Counterfactuals \cite{lewis2013counterfactuals}, as one of the popular methods using example-based explanations, adopt the thinking of what could be done differently in order to get a different outcome. Accordingly, counterfactuals focus on locally explaining the changing in the outcome due to changing details of input. Contrastive Explanation Method (CEM) \cite{dhurandhar2018explanations} is another method that seeks for explanations why predictions differ from another. Prototypes/Criticisms \cite{stock2018convnets} is another strategy that attempts to identify samples in data that are highly representative or not, respectively. In addition, KNN \cite{chakraborty2020making} and Trust Score \cite{druce2021explainable} can be also applied for local predictions.

\subsubsection{Perturbation-based Explanation}
Perturbation-based methods explain a black-box model by iteratively probing a with different variations of the inputs. Annotating, blurring, scrolling, and masking are some examples of distortions. It can be done at the feature level by substituting zero or random counterfactual samples for particular features, or by grouping a collection of pixels (superpixels) \cite{das2020opportunities}. Occlusion \cite{li2016understanding} is a simple local method by perturbing an instance in which the input features of an instance are systematically replaced with a constant value, usually zero. Also, Random Input Sampling for Explanations (RISE) \cite{petsiuk2018rise} probes a model with randomly masked portions of the input instance as a generalized kind of occlusion. As a classifier explanation method, Prediction Difference Analysis (PDA) \cite{zintgraf2017visualizing} assigns a significance value to each feature concerning each class and calculates the importance of a feature by observing how prediction changes when the value of a feature is uncertain. Occlusion is a local approach for perturbing an instance in which the input features of an instance are systematically substituted with a constant, generally zero. Meaningful Perturbation (MP) \cite{fong2017interpretable} is a local explanation approach for neural classifier predictions based on a framework of meta-predictors. These meta-predictors have been trained to predict whether input features are present or not. Their prediction error is a metric for how accurate the explanation is.

\subsubsection{Gradient/Backpropagation}
Instead of perturbation-based methods that focus on variations of inputs, gradient methods focus on the information flow. They use the information flow during backpropagation to determine the relationship between input features and output. Gradient-based methods often use heatmaps of neurons or feature attributions to provide visual explanations. Activation Maximization \cite{erhan2010understanding} is one of the examples of gradient methods that use also visual explanations. Saliency Maps \cite{simonyan2014deep} and DeepLift \cite{shrikumar2017learning} can be applied for local explanations. Also, DeepSHAP \cite{chen2021explaining} exploits compositional architecture to provide computational efficiency by extending KernelSHAP. Also, Deconvolution \cite{zeiler2014visualizing} and Guided Backpropagation \cite{springenberg2014striving} are generally used for explaining CNN-based black-box models. Class activation maps (CAM) \cite{zhou2016learning} is a model-specific explanation model used for CNN. Then, CAM is generalized with a new method called Gradient-weighed CAM (GradCAM) \cite{selvaraju2017grad} to support for more CNN architectures.

\section{Using XAI in IoT}
\label{Sec3_xai_inIoT}

The need for XAI is a debatable issue in the literature. According to Doshi-Velez \cite{doshi2017towards}, the explanations may not be necessary in some cases, and the system can be trusted (i) when the cost of explainability outweighs the need for it, (ii) when the impact of incorrect results on the field of application is not too severe, and (iii) when the problem has been thoroughly researched and applied to real-world scenarios. On the other hand, XAI is required if users are to understand, adequately trust, and effectively manage AI findings, whether, for commercial reasons, ethical problems, or regulatory concerns \cite{adadi2018peeking}.

The needs and benefits of adopting XAI approaches in the field of IoT are examined in this section. 
\begin{itemize}
    \item In crucial ML problems where not only the outcome but also the causes leading to the result are important, XAI approaches provide further information about why and how the prediction result is generated. For example, if an AI-driven aerial device employed in mission-critical applications strikes the wrong target, it is seen as a problem for the decision system. However, in black-box models, why the artificial intelligence behind the device makes such a judgment cannot be questioned or fully understood. With XAI, the possibilities causing the behavior of aerial devices can be analyzed in more detail and it becomes much easier and more understandable to find the cause of the error.
    \item It examines and regulates the behavioral patterns of machine learning models. For example, the data collected from IoT sensors often contain incomplete or erroneous data and may cause inaccurate decisions. XAI helps to discover and repair any unexpected faults or weaknesses, ensuring a smooth decision-making process. 
    \item Unlike the global outcomes of ML approaches, it can make local decisions about how it obtains the result of a single sample, thanks to its local interpretability ability. For example, in a healthcare system, an explanation is provided about a single patient in addition to the explanation extracted from all patients.
    \item It enables the discovery of hidden facts and information, as well as new perspectives on ML problems. If we can interpret the algorithm, we may discover important additional information and patterns that have not been noticed before.
    \item Unexpected deviations may occur in the training set of the ML algorithm. For example, it can lead to bias issues in a loan application approval algorithm, such as completely rejecting applications from a particular location, group, or race. XAI provides a guideline for revealing these biases in the model.
\end{itemize}

\section{XAI in IoT Application Domains}
\label{Sec4_xai_in_IoT_Domains}
In this section, we aim to provide a comprehensive overview/review of XAI studies in IoT application domains. For this purpose, we review the studies presented in the literature and summarize them in Table \ref{tab:DomainSpecific} in accordance with the XAI terminology and taxonomy.

\subsection{Autonomous Systems and Robotics}
In autonomous and robotics domain, AI-powered robots can perform tasks in critical and dangerous environments that humans cannot perform, or typical applications such as welding, ironing, painting, equipment placing/receiving and palletizing, all of which are performed with high strength, speed and accuracy \cite{ahmed2022artificial}. On the other hand, XAI algorithms make robots' reasoning explainable by humans, providing a better understanding of decisions and increasing the level of trust in robots.
Wang et al. \cite{wang2016impact} developed a new mechanism for robots to automatically generate explanations of their decisions based on Partially Observable Markov Decision Problems (POMDPs). They measured the performance of this mechanism in an agent-based test environment that simulates the tasks of a human-robot team. Experimental results show that robot explanations can improve task performance and improve trust and transparency. For robot explainability, the authors created a four-level natural language template including no explanation, explanation of two sensor readings, explanation of three sensor readings, and confidence-level explanation. In this way, robots are enabled to produce text explanations based on postdoc techniques related to their decision processes.

 Iyer et al. \cite{iyer2018transparency} proposed an explainable and object-sensitive deep reinforcement learning (DRL) model for the object recognition and classification problem. In the proposed model, the authors used a method that focuses on "object saliency maps" to provide human-intelligible visualization of DRLs states and actions. Kampik et al. \cite{kampik2019explaining} used text-based explainability approaches to explain the human-like actions of autonomous agents in human-robot interaction scenarios. The authors demonstrate the impact of the developed approaches on human participants through a study based on human-robot ultimatum games.

Guo \cite{guo2020partially} proposed the Double Dueling Deep Q-learning Neural Network (DDDQN) network for the quality of service (QoS), Quality-of-Experience (QoE), and energy optimization of the UAVs in the UAV-supported 5G network. In the study, partial explainability is provided by extracting the state weights of the proposed DDDQN network.

\subsection{Energy Management}
AI-powered energy decision systems used in smart cities, smart grids and smart home applications provide decision makers with critical predictions about energy use and emissions. However, decision-makers want to know and believe the parameters that affect the model outputs in addition to the predicted results. 
Kim and Cho \cite{kim2019electric} proposed an energy demand prediction model for smart environments. Authors use an explainable autoencoder-based deep learning model to predict energy demands in various environment states. The proposed model produces consumption estimates of 15, 30, and 45 minutes according to defined states in the environment. The developed model has been tested on home electrical energy consumption data set containing five-year data. The model predicts energy demands by taking the 8 features (date, global active power, global reactive power, global intensity, voltage, sub-metering 1/2/3) in the environment as inputs. The authors provide more explainable information by visualizing monthly electricity demands with t-SNA algorithm. Experimental results show that the proposed model achieves the best performance compared to other machine learning models (Decision Trees, MLP, LSTM etc.).
Sirmacek and Riverio \cite{sirmacek2020occupancy} proposed two new algorithms (machine learning-based, computer vision-based) to predict real-time occupancy in smart office spaces by using very low-resolution heat sensors data. The authors used the well-known approach SHapley Additive exPlanations (SHAP) to extract the contribution of features to the classification in the first algorithm and to reveal the contribution of each local pixel in the second algorithm. They showed that proposed algorithms can be used in many application areas related to the automation and efficient use of offices, spaces, and buildings.
Amiri et al. \cite{amiri2021peeking} proposed an artificial neural network-based transportation energy model to predict urban transportation energy. They used LIME for model transparency and explainability.

\subsection{Environmental Monitoring}
Environmental applications in IoT focus on many issues and problems in areas such as smart cities, air monitoring, water management and climate monitoring, etc. In solving the addressed problems, mostly AI models with high complexity and low interpretability are used. Therefore, interpretable and more transparent models are needed in these areas that affect human and environmental health. In this context, there are some current efforts to solve the existing problems. Barrett-Powell et al. \cite{barrett2020experimentation} proposed a Situational Understanding Explorer (SUE) platform for coalition situational understanding research that highlights capabilities in explainable artificial intelligence for event processing in a dense urban terrain setting. 
Kalamaras et al. \cite{kalamaras2019visual} proposed a new visual analytics platform that includes architecture and components created for air pollution monitoring within the scope of the AI4IoT project. In this context, there are two component designs specifically designed to provide an explanation of artificial intelligence models. The first is the Annotated Line Chart, which visualizes the parameters of an ARIMA prediction model, the second is the SHAP Chart provides insight into the most important features used for Random Forest regression.
Graham et al. \cite{graham2020genome} developed an environmental bio-sensing platform called Dynomics to evaluate patterns in transcriptional data at a genome-scale by using XGBoost, LSTM classifiers and XAI algorithms. They used SHAP algorithm to gain insight into the features used by the predictive algorithms trained on transcriptional data.
Thakker et al. \cite{thakker2020explainable} proposed a flood monitoring application using semantic web technologies for the flood problem in smart cities. In this application, the authors created an explainable hybrid image classification model by combining CNN-based DL models and semantic techniques. CNN was used to determine the object coverage proportion in the drainage and gully images obtained from critical geographic areas, whereas semantic techniques have been used to define the relationship between the coverage level and the objects. In this work, the authors preferred rule-based explainability in image classification. Ryo et al. \cite{ryo2021explainable} performed an XAI-based animal SDM (species distribution model) analysis over time in terms of ecology, biogeography and conservation biology. In the work, the authors demonstrated that XAI can be used to improve the interpretability of SDMs by performing the distribution analysis of African elephants with the LIME model. 
Daillo et al. \cite{diallo2020explainable} proposed a CNN algorithm that aims to reduce the problem of adaptation space and can be used in an on-campus smart environment monitoring platform. The authors aim to design a reliable system using explainable artificial intelligence in the learning and prediction process of the proposed algorithm.

\begin{table*}[htb]
\centering
\caption{The Summary of XAI Studies by IoT Application Domain}
\label{tab:DomainSpecific}
\resizebox{\textwidth}{!}{%
\begin{tabular}{|l|l|l|l|l|lll|}
\hline
\multirow{2}{*}{\textbf{IoT Domain}} & \multirow{2}{*}{\textbf{Reference}} & \multirow{2}{*}{\textbf{Year}} & \multirow{2}{*}{\textbf{ML/DL Model(s)}} & \multirow{2}{*}{\textbf{XAI Model Name}} & \multicolumn{3}{c|}{\textbf{XAI Taxonomy}} \\ \cline{6-8} 
 &  &  &  &  & \multicolumn{1}{l|}{\textbf{\begin{tabular}[c]{@{}l@{}}Ante-hoc/\\ Post-Hoc\end{tabular}}} & \multicolumn{1}{l|}{\textbf{\begin{tabular}[c]{@{}l@{}}Model Spesific /\\ Model Agnostic\end{tabular}}} & \textbf{\begin{tabular}[c]{@{}l@{}}Local /\\ Global\end{tabular}} \\ \hline
\multirow{2}{*}{\begin{tabular}[c]{@{}l@{}}Autonomous Systems \\ and Robotics\end{tabular}} & Iyer et al. \cite{iyer2018transparency}& 2018 & DRLN & Saliency Map & \multicolumn{1}{l|}{Post-hoc} & \multicolumn{1}{l|}{Agnostic} & Local \\ \cline{2-8} 
 & Guo et al. \cite{guo2020partially}& 2020 & DRL & - & \multicolumn{1}{l|}{Ante-hoc} & \multicolumn{1}{l|}{Specific} & Local \\ \hline
\multirow{3}{*}{Energy Management} & Kim et al. \cite{kim2019electric}& 2019 & Autoencoder & t-SNE & \multicolumn{1}{l|}{Post-hoc} & \multicolumn{1}{l|}{Agnostic} & - \\ \cline{2-8} 
 & Sirmacek et al. \cite{sirmacek2020occupancy}& 2020 & CatBoost & SHAP & \multicolumn{1}{l|}{Ante-hoc} & \multicolumn{1}{l|}{Agnostic} & Local \\ \cline{2-8} 
 & Amiri et al. \cite{amiri2021peeking}& 2021 & NN & LIME & \multicolumn{1}{l|}{Post-hoc} & \multicolumn{1}{l|}{Agnostic} & Local \\ \hline
\multirow{4}{*}{Environment} & Kalamaras et al. \cite{kalamaras2019visual}& 2019 & ARIMA, RF & SHAP & \multicolumn{1}{l|}{Post-hoc} & \multicolumn{1}{l|}{Agnostic} & Local \\ \cline{2-8} 
 & Diallo et al. \cite{diallo2020explainable}& 2020 & CNN & Integrated Gradients & \multicolumn{1}{l|}{Post-hoc} & \multicolumn{1}{l|}{Agnostic} & Local \\ \cline{2-8} 
 & Graham et al. \cite{graham2020genome}& 2020 & LSTM, XGBoost & SHAP & \multicolumn{1}{l|}{Post-hoc} & \multicolumn{1}{l|}{Agnostic} & Local \\ \cline{2-8} 
 & Ryo et al. \cite{ryo2021explainable}& 2021 & SDM & LIME & \multicolumn{1}{l|}{Post-hoc} & \multicolumn{1}{l|}{Agnostic} & Local \\ \hline
\multirow{4}{*}{Finance} & Sachan et al. \cite{sachan2020explainable}& 2019 & - & Belief-Rule-Base & \multicolumn{1}{l|}{-} & \multicolumn{1}{l|}{-} & - \\ \cline{2-8} 
 & Bussman et al. \cite{bussmann2021explainable}& 2020 & XGBoost & Shapley Values & \multicolumn{1}{l|}{Post-hoc} & \multicolumn{1}{l|}{Agnostic} & Global \\ \cline{2-8} 
 & Gramegna et al. \cite{gramegna2020buy}& 2020 & XGBoost & SHAP & \multicolumn{1}{l|}{Post-hoc} & \multicolumn{1}{l|}{Agnostic} & Local \\ \cline{2-8} 
 & Gite et al. \cite{gite2021explainable}& 2021 & LSTM-CNN & LIME & \multicolumn{1}{l|}{Post-hoc} & \multicolumn{1}{l|}{Agnostic} & Local \\ \hline
\multirow{10}{*}{Healthcare} & Chittajallu et al. \cite{chittajallu2019xai}& 2019 & ResNet50 & XAI-CBIR & \multicolumn{1}{l|}{Post-hoc} & \multicolumn{1}{l|}{Agnostic} & - \\ \cline{2-8} 
 & Hossain et al. \cite{hossain2020explainable}& 2020 & \begin{tabular}[c]{@{}l@{}}ResNet50, \\ Deep tree, \\ Inception v3\end{tabular} & LIME, GradCAM & \multicolumn{1}{l|}{Post-hoc} & \multicolumn{1}{l|}{Agnostic} & - \\ \cline{2-8} 
 & Monroe et al. \cite{monroe2021hiho}& 2020 & CNN & HihO & \multicolumn{1}{l|}{Post-hoc} & \multicolumn{1}{l|}{Agnostic} & Global \\ \cline{2-8} 
 & Pnevmatikakis et al. \cite{pnevmatikakis2021risk}& 2020 & RF, DNN & SHAP & \multicolumn{1}{l|}{Post-hoc} & \multicolumn{1}{l|}{Agnostic} & Local \\ \cline{2-8} 
 & Hatwell et al. \cite{hatwell2020ada}& 2020 & AdaBoost & Ada-WHIPS & \multicolumn{1}{l|}{Post-hoc} & \multicolumn{1}{l|}{Agnostic} & Local \\ \cline{2-8} 
 & Dave et al. \cite{dave2020explainable}& 2020 & XGBoost & LIME, SHAP & \multicolumn{1}{l|}{Post-hoc} & \multicolumn{1}{l|}{Agnostic} & \begin{tabular}[c]{@{}l@{}}Local, \\ Global\end{tabular} \\ \cline{2-8} 
 & Gozzi et al. \cite{gozzi2022xai}& 2022 & CNN & Grad-CAM, SHAP & \multicolumn{1}{l|}{Post-hoc} & \multicolumn{1}{l|}{Specific} & Global \\ \hline
\multirow{9}{*}{Industrial} & Rehse et al. \cite{rehse2019towards}& 2019 & DNN & - & \multicolumn{1}{l|}{Post-hoc} & \multicolumn{1}{l|}{Agnostic} & \begin{tabular}[c]{@{}l@{}}Local, \\ Global\end{tabular} \\ \cline{2-8} 
 & Chen and Lee \cite{chen2020vibration}& 2020 & CNN & Grad-CAM & \multicolumn{1}{l|}{Post-hoc} & \multicolumn{1}{l|}{Agnostic} & Local \\ \cline{2-8} 
 & Sun et al. \cite{sun2020vision}& 2020 & CNN & CAM & \multicolumn{1}{l|}{Post-hoc} & \multicolumn{1}{l|}{Agnostic} & Local \\ \cline{2-8} 
 & Serradilla et al. \cite{serradilla2020interpreting}& 2020 & RF & LIME, ELI5 & \multicolumn{1}{l|}{Post-hoc} & \multicolumn{1}{l|}{Agnostic} & \begin{tabular}[c]{@{}l@{}}Local,  \\ Global\end{tabular} \\ \cline{2-8} 
 & Senoner et al. \cite{senoner2021using}& 2021 & DT & SHAP & \multicolumn{1}{l|}{Post-hoc} & \multicolumn{1}{l|}{Agnostic} & Local \\ \cline{2-8} 
 & Mehdiyev et al. \cite{mehdiyev2021explainable}& 2021 & DNN & \begin{tabular}[c]{@{}l@{}}Surrogate \\ Decision Trees\end{tabular} & \multicolumn{1}{l|}{Post-hoc} & \multicolumn{1}{l|}{Agnostic} & Local \\ \cline{2-8} 
 & Brito et al. \cite{brito2022explainable}& 2022 & kNN, CBLOF & SHAP & \multicolumn{1}{l|}{Post-hoc} & \multicolumn{1}{l|}{Agnostic} & Local \\ \hline
\multirow{8}{*}{Security and Privacy} & Saharkhizan et al. \cite{saharkhizan2020ensemble}& 2020 & LSTM & DT & \multicolumn{1}{l|}{Post-hoc} & \multicolumn{1}{l|}{Specific} & Global \\ \cline{2-8} 
 & Khan et al. \cite{khan2021new}& 2021 & CNN, AE-LSTM & LIME & \multicolumn{1}{l|}{Post-hoc} & \multicolumn{1}{l|}{Agnostic} & Local \\ \cline{2-8} 
 & Sarhan et al. \cite{sarhan2021explainable}& 2021 & DFF, RF & SHAP & \multicolumn{1}{l|}{Post-hoc} & \multicolumn{1}{l|}{Agnostic} & Local \\ \cline{2-8} 
 & Mahbooba et al. \cite{mahbooba2021explainable}& 2021 & DT & - & \multicolumn{1}{l|}{Ante-hoc} & \multicolumn{1}{l|}{Specific} & Global \\ \cline{2-8} 
 & Nascita et al. \cite{nascita}& 2021 & BiGRU & DeepSHAP & \multicolumn{1}{l|}{Post-hoc} & \multicolumn{1}{l|}{Agnostic} & Global \\ \cline{2-8} 
 & Zolanvari et al. \cite{zolanvari2021trust}& 2021 & ANN & TRUST & \multicolumn{1}{l|}{Ante-hoc} & \multicolumn{1}{l|}{Agnostic} & Local \\ \cline{2-8} 
 & Le et al. \cite{le2022classification}& 2022 & DT, RF & SHAP & \multicolumn{1}{l|}{Post-hoc} & \multicolumn{1}{l|}{Agnostic} & \begin{tabular}[c]{@{}l@{}}Local, \\ Global\end{tabular} \\ \hline
\multirow{4}{*}{Smart Agriculture} & Kundu et al. \cite{kundu2021iot}& 2021 & Custom-Net & Grad-CAM & \multicolumn{1}{l|}{Post-hoc} & \multicolumn{1}{l|}{Agnostic} & Local \\ \cline{2-8} 
 & Viana et al. \cite{viana2021evaluation}& 2021 & RF & LIME, PDP & \multicolumn{1}{l|}{Post-hoc} & \multicolumn{1}{l|}{Agnostic} & \begin{tabular}[c]{@{}l@{}}Local, \\ Global\end{tabular} \\ \cline{2-8} 
 & Garrido et al. \cite{garrido2022evaporation}& 2022 & ANN & DT & \multicolumn{1}{l|}{Ante-hoc} & \multicolumn{1}{l|}{Specific} & Global \\ \hline
\end{tabular}%
}
\end{table*}
\subsection{Financial System}
AI models are rapidly changing the way the financial system works, providing cost savings as well as operational efficiency in areas such as asset management, investment advisory, risk forecasting, lending and customer service \cite{danielsson2021artificial}. However, due to the nature of financial domain, AI decisions in these applications also contain risks that can be costly due to their consequences. For this reason, XAI should be considered among the priority issues in the financial field. Sachan et al. \cite{sachan2020explainable} developed a belief-rule-based (BRB) explainable decision support system to automate the process of lending loans. The authors aimed to reveal the chain of events that clarified the decision process by adding a structure containing factual and heuristic rules to the traditional IF-THEN rule-based system. Bussmann et al. \cite{bussmann2021explainable} proposed an XAI model that can be used to measure the risks arising in loan purchases. The proposed model aimed to design a system that can explain the credit score of borrowers and predict their future behavior. The explainability of the proposed model is provided by using the TreeSHAP method, which provides explanations based on Shapley values. 
Gramegna and Giudici \cite{gramegna2020buy} proposed a technological insurance model for the insurance industry that allows understanding of purchase and cancellation behavior of customers. The authors used the XGBoost algorithm for the extraction of the behavior patterns and SHAP for the model agnostic interpretability.
Gite et al. \cite{gite2021explainable} proposed a stock price prediction moded based on LSTM and CNN. With the proposed model, it is aimed that investors can learn how and when stock prices fall or rise and make decisions accordingly. In this study, the interpretation and explainability of the model outputs were carried out using LIME.

\subsection{Healthcare}
Decision-making in the healthcare domain affect people directly, and it is very difficult to compensate for their negative consequences. Therefore, AI models used in this field should not only perform well but also be reliable, transparent, interpretable and explainable. Especially in IoT applications that provide monitoring, diagnosis and health advice, this need should be addressed as a priority. There are many studies for this purpose. For example,
Chittajallu et al. \cite{chittajallu2019xai} proposed a human-assisted explainable AI system called XAI-CBIR that enables content-based image retrieval for use in surgical education. In XAI-CBIR, CNN-based DL model lists similar pictures by extracting the semantic descriptors of the image in the query video. It iteratively trains itself based on relevant feedback from users. The developed system provides the explainability of the pictures similar to the picture in the query by creating a visual saliency map.
Hossain et al. \cite{hossain2020explainable} proposed a three-tiered (stakeholder layer, edge layer, and cloud layer) smart healthcare framework that uses 5G networking to combat COVID-19-like pandemics. The framework is capable of detecting COVID-19 using chest X-ray or CT scan images, as well as features such as social distancing, masks and body temperature control. The authors used ResNet50, Deep tree, and Inception v3 models in the edge layer.
Interactive explainability is provided by using Local Interpretable Model-Agnostic (LIMA) on the knowledge graphs produced based on the learning parameters of these DL models.
In \cite{dave2020explainable}, Dave et al. focused on the usability of feature- and example-based XAI techniques on the heart disease dataset to ensure the reliability of AI systems used in the healthcare domain. The authors showed that black box model behaviors can be explained using feature-based XAI techniques SHAP and LIME, and example-based techniques Anchors, Counterfactuals, Integrated gradients, CEM, KernelSHAP.
Hatwell et al. \cite{hatwell2020ada} developed a new Adaptive Weighted High-importance Path Particles (Ada-WHIPS) model to make the AdaBoost model, which uses computer-assisted diagnostics in healthcare, more explainable. Ada-WHISP uses a new formulation to explain the classification of AdaBoost models with simple classification rules. Pnevmatikakis et al. \cite{pnevmatikakis2021risk} developed a risk assessment system for professionals in the health insurance sector. The proposed system also includes a virtual coaching system that provides the prediction of people's lifestyles and the production of applicable lifestyle recommendations. In this system, RandomForest and DNN algorithms are used to predict the lifestyle of individuals. SHAP was used for the explainability of the prediction results.
Monroe et al. \cite{monroe2021hiho} proposed a CNN-based hierarchical occlusion (HihO) model that rapidly increases the interpretability of statistical findings in medical imaging workflows in IoT healthcare applications. The authors compared the developed method with GradCAM and (Parkinson Progression Markers Initiative) RISE methods on the Parkinson's Progression Markers Initiative (PPMI) dataset. The proposed model has been shown to render 20 and 200 times faster than GradCAM and RISE models, respectively. Gozzi et al. \cite{gozzi2022xai} focused on the explainability of IA models used to classify hand movements based on EMG signals. The authors specifically investigated the effect of XAI models to provide improvements in the life of amputees using myo-controlled prostheses. The authors used SVM, LDA, XRT and CNN algorithms in the classification process of hand gestures, and gradCAM and SHAP in the XAI process.

\subsection{Industrial Domain}
With industry 4.0, AI systems in IIoT enable machines to perform tasks such as self-monitoring, interpretation, diagnosis and analysis autonomously in production lines and processes of manufacturing, logistics and related industries. In IIoT, XAI methods can enable the adoption of new technologies and digital transformation, and better product quality controls.
Oyekunlu \cite{oyekanlu2018distributed} developed LSTM-RNN-based explainable deep learning models for the prediction of time series data based on energy and electricity consumption in Industrial IoT systems. Later, he designed an IIoT system based on edge, fog and cloud layers for the applicability of these models. Distributed osmotic computing approach is used to show how low-cost hardware can be used in the designed system. In the proposed IIoT system, data preprocessing and data quality enhancement in edge devices, and developped LSTM and RNN models are used in the fog layer. Christou et al. \cite{christou2020predictive} focused on estimating the Remaining Useful Life (RULs) of machines on production lines using the Qarma family of algorithms, which provides rule-based explainability for industry IoT applications. The authors predict the RULs of the drilling machine and the Quality Management in the Wheel Production Line in the Automotive industry based on rule-based explainability.
Rehse et al. \cite{rehse2019towards} focused on real-time process management in the smart lego factory for AI-based Industry 4.0 applications. In this context, they developed an RNN-based deep learning model that performs process predictions. In order to make the process outcome estimates more understandable to workers and visitors, they designed an interface that offers global and local explanations from poct-hoc explainability techniques.
\par In Industrial IoT, unforeseen failures can cause disruption of production processes, jeopardize employee safety and increase operational costs. Therefore, it is critical to monitor the health status of the mechanical equipment and to diagnose the failure conditions. Sun et al. \cite{sun2020vision} proposed a Convolutional Neural Network (CNN)-based DL model for machine health monitoring and automatic fault diagnosis. The authors have integrated a layer called Class Activation Maps (CAMs) into the proposed model, which provides a visual explanation. In this way, it provides a visual explanation of the image by localizing the damaged part with CAM without placing any sensors on the machines.
Chen and Lee \cite{chen2020vibration} focused on the development of explicable CNNs for classification in vibration signals analysis. The authors used Gradient class activation mapping (Grad-CAM), which generates heat maps by calculating the weights of each feature map according to the classification scores, for CNN explainability. The authors also verified the model explainability with NN, ANFIS and decision trees.

Serradilla et al. \cite{serradilla2020interpreting} developed an RF algorithm for estimating the remaining useful life of industrial machines. ELI5 and LIME techniques were used for the local and global interpretability of the created model.
Senoner et al. \cite{senoner2021using} designed a DT-based decision model to improve process quality in the manufacturing process. The designed model was tested on the transistor chip production line. The SHAP model was used in the analysis of the relationship between the parameters in the production and the quality of the production process. Mehdiyev and Fettke \cite{mehdiyev2021explainable} developed a conceptual framework for predictive process monitoring that provides approaches to guide researchers and practitioners. In this context, they proposed a new post-doc explanation approach called Surrogate Decision Trees, which will enable the results of models such as DNN, CNN, LSTM and GAN to be understood. 
Brito et al. \cite{brito2022explainable} developed an approach for fault detection and detection in rotating machines based on many unsupervised anomaly detections and clustering algorithms such as KNN, SVM, Histogram-based outlier score (HBOS), Isolation forest (IF), and Local outlier factor (LOF). The explainability of the models used in this approach is provided by SHAP and Local Depth-based Feature Importance for the Isolation Forest (Local-DIFFI). The proposed approach has been tested on the bearing, chance, and mechanical failure datasets.
\subsection{Security and Privacy}
Security and privacy are of paramount importance in all applications of IoT. AI and ML-based algorithms/models are widely used for intrusion detection, anomalous traffic detection, authentication and malware detection in IoT security systems. However, these systems need XAI techniques for interpreting model decisions, reasoning and providing trust management.
\begin{landscape}
\begin{table}[]
\centering
\caption{THe summary of XAI Studies According to Their Methodologies}
 
\label{tab:MethodSpecific}
\resizebox{0.94\linewidth}{!}{%
\begin{tabular}{|l|l|l|l|ll|l|l|}
\hline
\multirow{2}{*}{\textbf{\begin{tabular}[c]{@{}l@{}}Explanation Type/\\ Method\end{tabular}}} & \multirow{2}{*}{\textbf{Reference}} & \multirow{2}{*}{\textbf{Problem Type}} & \multirow{2}{*}{\textbf{ML/DL Model(s)}} & \multicolumn{2}{c|}{\textbf{Data Type}} & \multirow{2}{*}{\textbf{Dataset}} & \multirow{2}{*}{\textbf{Data Provider}} \\ \cline{5-6}
 &  &  &  & \multicolumn{1}{l|}{Input Data} & Output Data &  &  \\ \hline
Example-based & Zolanvari et al. \cite{zolanvari2021trust}& Classification & ANN & \multicolumn{1}{l|}{Numerical} & Numerical & WUSTL-IIoT & - \\ \hline
\multirow{28}{*}{Feature Importance} & Guo et al. \cite{guo2020partially}& - & DRL & \multicolumn{1}{l|}{Pictorial} & Numerical & \begin{tabular}[c]{@{}l@{}}Base station data from London, \\ Geo-tagged tweets\end{tabular} & - \\ \cline{2-8} 
 & Sirmacek et al. \cite{sirmacek2020occupancy}& Classification & CatBoost & \multicolumn{1}{l|}{Pictorial} & Numerical & Heat sensor data & - \\ \cline{2-8} 
 & Kalamaras et al. \cite{kalamaras2019visual}& Regression & ARIMA, RF & \multicolumn{1}{l|}{Time-series} & Numerical & Pollution, weather, and traffic data & \begin{tabular}[c]{@{}l@{}}Norwegian Environment Agency\\ Norwegian Meteorological Institute\\ Norwegian Road Authorities\end{tabular} \\ \cline{2-8} 
 & Ryo et al. \cite{ryo2021explainable}& Regression & SDM & \multicolumn{1}{l|}{Numerical} & Numerical & - & Zenodo Digital Repository \\ \cline{2-8} 
 & Graham et al. \cite{graham2020genome}& Classification & LSTM, XGBoost & \multicolumn{1}{l|}{Numerical} & Numerical & Dynomics data & \begin{tabular}[c]{@{}l@{}}University of California San Diego \\ Biodynamics Laboratory\end{tabular} \\ \cline{2-8} 
 & Bussman et al. \cite{bussmann2021explainable}& Classification & XGBoost & \multicolumn{1}{l|}{Numerical} & Numerical & Credit scoring data & European External Credit Assessment Institution (ECAI) \\ \cline{2-8} 
 & Pnevmatikakis et al. \cite{pnevmatikakis2021risk}& Classification & RF, DNN & \multicolumn{1}{l|}{Categorical} & Numerical & RWD & Samsung Health \\ \cline{2-8} 
 & Rehse et al. \cite{rehse2019towards}& Classification & DNN & \multicolumn{1}{l|}{\begin{tabular}[c]{@{}l@{}}Numerical \& \\ Categorical\end{tabular}} & Textual & Sensor and production data & The DFKI-Smart-Lego-Factory \\ \cline{2-8} 
 & Gramegna et al. \cite{gramegna2020buy}& Classification & XGBoost & \multicolumn{1}{l|}{\begin{tabular}[c]{@{}l@{}}Numerical \& \\ Categorical\end{tabular}} & Numerical & - & - \\ \cline{2-8} 
 & Senoner et al. \cite{senoner2021using}& Regression & DT & \multicolumn{1}{l|}{Numerical} & Numerical & Transistor chip production data & HitachiABB \\ \cline{2-8} 
 & Serradilla et al. \cite{serradilla2020interpreting}& Regression & RF & \multicolumn{1}{l|}{Time-series} & Numerical & - & - \\ \cline{2-8} 
 & Brito et al. \cite{brito2022explainable}& Classification & kNN and CBLOF & \multicolumn{1}{l|}{Numerical} & Numerical & \begin{tabular}[c]{@{}l@{}}Bearing dataset, Gearbox dataset, \\ Mechanical fault dataset\end{tabular} & - \\ \cline{2-8} 
 & Sarhan et al. \cite{sarhan2021explainable}& Classification & DFF,   RF & \multicolumn{1}{l|}{Numerical} & Numerical & CSE-CIC-IDS2018, BoT-IoT,  ToN-IoT & \begin{tabular}[c]{@{}l@{}}University of New Brunswick,\\ Intelligent Security Group UNSW Canberra,\\ Australian Defence Force Academy (ADFA)\end{tabular} \\ \cline{2-8} 
 & Amiri et al. \cite{amiri2021peeking}& Classification & NN & \multicolumn{1}{l|}{\begin{tabular}[c]{@{}l@{}}Numerical \& \\ Categorical\end{tabular}} & Numerical & Household Travel Survey (HTS) data & - \\ \cline{2-8} 
 & Le et al. \cite{le2022classification}& Classification & DT,   RF & \multicolumn{1}{l|}{\begin{tabular}[c]{@{}l@{}}Numerical \& \\ Categorical\end{tabular}} & Visual & \begin{tabular}[c]{@{}l@{}}NF-BoT-IoT-v2,  NF-ToN-IoT-v2, \\ IoTDS20\end{tabular} & - \\ \cline{2-8} 
 & Kundu et al. \cite{kundu2021iot}& Classification & Custom-Net & \multicolumn{1}{l|}{Pictorial} & Visual & Imagery and Parametric data & \begin{tabular}[c]{@{}l@{}}Indian Council of Agricultural Research (ICAR),\\ All India Coordinated Research Project (AICRP)\end{tabular} \\ \cline{2-8} 
 & Viana et al. \cite{viana2021evaluation}& Classification & RF & \multicolumn{1}{l|}{Numerical} & Visual & Agricultural Parcel-data & - \\ \cline{2-8} 
 & Dave et al. \cite{dave2020explainable}& Classification & XGBoost & \multicolumn{1}{l|}{Numerical} & Numerical & Heart Disease Dataset & UCI Machine Learning Repository \\ \hline
\multirow{2}{*}{\begin{tabular}[c]{@{}l@{}}Feature Importance and\\ Perturbation\end{tabular}} & Gite et al. \cite{gite2021explainable}& Classification & LSTM-CNN & \multicolumn{1}{l|}{Textual} & Visual & \begin{tabular}[c]{@{}l@{}}News Headlines dataset, \\ Yahoo Finance dataset\end{tabular} & Pulse,   Yahoo \\ \cline{2-8} 
 & Khan et al. \cite{khan2021new}& Classification & CNN,   AE-LSTM & \multicolumn{1}{l|}{Time-series} & Numerical & Real-world gas pipeline system data & - \\ \hline
\begin{tabular}[c]{@{}l@{}}Feature Importance and\\ Gradient/Backpropagation\end{tabular} & Gozzi et al. \cite{gozzi2022xai}& Classification & CNN & \multicolumn{1}{l|}{Pictorial} & Visual & EMG data & - \\ \hline
\multirow{6}{*}{Gradient/Backpropagation} & Iyer et al. \cite{iyer2018transparency}& Classification & DRLN & \multicolumn{1}{l|}{Pictorial} & Visual & MS. Pacmann game screenshots & - \\ \cline{2-8} 
 & Diallo et al. \cite{diallo2020explainable}& N/D & CNN & \multicolumn{1}{l|}{Time-series} & Numerical & Space Suttle Marotta valve time series dataset & - \\ \cline{2-8} 
 & Chen and Leeok \cite{chen2020vibration}& Classification & CNN & \multicolumn{1}{l|}{Pictorial} & Visual & Bearing dataset & Case Western Reserve University (CWRU) \\ \cline{2-8} 
 & Sun et al. \cite{sun2020vision}& Classification & CNN & \multicolumn{1}{l|}{Pictorial} & Visual & \begin{tabular}[c]{@{}l@{}}Base-excited cantilever beam dataset \\ and Water pump dataset\end{tabular} & - \\ \cline{2-8} 
 & Nascita et al. \cite{nascita}& Classification & BiGRU & \multicolumn{1}{l|}{Numerical} & Numerical & MIRAGE-2019 & - \\ \cline{2-8} 
 & Saharkhizan et al. \cite{saharkhizan2020ensemble}& Classification & LSTM & \multicolumn{1}{l|}{Time-series} & Numerical & - & - \\ \hline
\begin{tabular}[c]{@{}l@{}}Perturbation and\\ Visualization\end{tabular} & Hossain  et al. \cite{hossain2020explainable}& Classification & \begin{tabular}[c]{@{}l@{}}ResNet50,\\ Deep tree,\\ Inception v3\end{tabular} & \multicolumn{1}{l|}{Pictorial} & Visual & - & - \\ \hline
\multirow{4}{*}{Rule Extraction} & Hatwell et al. \cite{hatwell2020ada}& Classification & AdaBoost & \multicolumn{1}{l|}{\begin{tabular}[c]{@{}l@{}}Numerical \& \\ Categorical\end{tabular}} & Rules & \begin{tabular}[c]{@{}l@{}}Breast cancer, Cardiotocography, \\ Diabetic retinopathy, Cleveland heart, \\ Mental health survey 14/16, \\ Hospital readmission, Thyroid,\\ Understanding society\end{tabular} & UCI Machine Learning Repository \\ \cline{2-8} 
 & Mehdiyev et al. \cite{mehdiyev2021explainable}& Classification & DNN & \multicolumn{1}{l|}{\begin{tabular}[c]{@{}l@{}}Numerical \& \\ Categorical\end{tabular}} & Numerical & Real-life process log data & Volvo IT Belgium’s incident management system \\ \cline{2-8} 
 & Garrido et al. \cite{garrido2022evaporation}& Regression & ANN & \multicolumn{1}{l|}{Numerical} & Numerical & The climate data & - \\ \hline
\multirow{4}{*}{Transparent} & Sachan et al. \cite{sachan2020explainable}& Classification & N/D & \multicolumn{1}{l|}{Numerical} & Textual & Credit data & Credit bureau server \\ \cline{2-8} 
 & Mahbooba et al. \cite{mahbooba2021explainable}& Classification & DT & \multicolumn{1}{l|}{\begin{tabular}[c]{@{}l@{}}Numerical \& \\ Categorical\end{tabular}} & Rules & KDD benchmark dataset & \begin{tabular}[c]{@{}l@{}}The UCI KDD Archive Information and \\ Computer Science \\ University of California, Irvine\end{tabular} \\ \hline
\multirow{3}{*}{Visualization} & Kim et al. \cite{kim2019electric}& Regression & Autoencoder & \multicolumn{1}{l|}{Numerical} & Numerical & Household electric power consumption & UCI Machine Learning Repository \\ \cline{2-8} 
 & Chittajallu et al. \cite{chittajallu2019xai}& Classification & ResNet50 & \multicolumn{1}{l|}{Pictorial} & Visual & Cholec80 & - \\ \cline{2-8} 
 & Monroe et al. \cite{monroe2021hiho}& Classification & CNN & \multicolumn{1}{l|}{Pictorial} & Visual & PPMI RD & - \\ \hline
\end{tabular}%
}
\end{table}
\end{landscape}
Liu et al. \cite{liu2021zero} proposed a DNN-based framework to quickly and reliably detect time-dependent anomalous events in IoT data. Voronoi diagrams were used to explain the DNN classification model. The framework is tested on both real aviation communication systems and simulation data, and its effectiveness has been demonstrated.
Mahbooba et al. \cite{mahbooba2021explainable} considered the explainability of ML classifiers to improve trust management in intrusion detection systems. The study focused on the explainability method based on rule inference. In the study, they compared the accuracy and explainability of classifiers such as decision trees, random forest and SVM on the widely used KDD benchmark dataset. The results show that the decision tree classifier has both higher accuracy and produces more interpretable rules. For the same research problem, Sarhan et al. \cite{sarhan2021explainable} proposed an interpretable ML-based IDS to ensure security in IoT networks. The proposed system is capable of detecting different types of attacks in various network environments and has a generalizable structure. The explainability and interpretability of the ML algorithms in the proposed system were carried out through the SHAP method. In another study of the same scope, Le et al. \cite{le2022classification} aimed to increase the intrusion detection performance of IDS systems working with ML algorithms in IoT-based security system datasets. The authors used DT and RF algorithms and used SHAP to explain and interpret the classification decisions of these algorithms.

Zolanvari et al. \cite{zolanvari2021trust} developed a new XAI model called Transparency Based on Statistical Theory (TRUST) that statistically explains model outputs in AI-based systems. The authors used factor analysis in the transformation of the model inputs, and multi-modal gaussian distribution in determining the model output probabilities. The developed TRUST model was compared with LIME model and it was emphasized that it was more successful than LIME in terms of speed and accuracy. Khan et al. \cite{khan2021new} proposed a framework based on CNN and LSTM models for the detection of cyber threats in IIoT networks. In the proposed framework, complex attacks are examined on time series data with Sliding Window(SW) technique with fixed length and classified with DL models. The authors used LIME technique for DL models explainability. Similarly, in \cite{saharkhizan2020ensemble}, Saharkhizan et al designed multiple LSTM models for cyber-attack detection in IoT networks. then authors used a decision tree to unify and explain the outputs of LSTM models. Nascita et al. \cite{nascita} designed an architecture called MIMETIC-ENHANCED, which can perform traffic classification of mobile IoT devices based on explainability analysis. In this architecture, the authors used the bidirectional GRU model for traffic classification within the architecture and the DeepSHAP method for the explainability of the BiGRU model.
\subsection{Smart Agriculture}
Smart agriculture opened the door to sustainable and responsible agriculture by making better management decisions with AI-powered decision support systems. However, these systems face user adoption issues. For this reason, it is critical to make AI systems used in agriculture interpretable and understandable by the user. In this context, there are some studies in the literature. Tsakiridis et al. \cite{tsakiridis2020versatile} propose a decision support system called Vital that fully automates the irrigation of open fields. In the proposed system, the authors use the explainable AI approach to show that the Vital can help conserve water and resources by making more precise decisions in irrigation management.
Gandhi et al. \cite{gandhi2021explainable} propose a framework that allows for field water system mechanization, irrigation control and precision farming based on a fuzzy logic approach. Authors aim to obtain the exact and the best atmosphere where a crop can easily cultivate and can give maximum yield. In this context, the proposed model is tested with different crops (barley, cotton, millets, groundnut, etc.) and the soil types (clayey, red, sandy, etc.). For this purpose, the proposed Mamdani-type rule-based system takes temperature, humidity, soil moisture, and water nutrient sensor values, then it makes smart decisions. 
Kenny et al. \cite{kenny2021bayesian} proposed a case-based reasoning system called PBI-CBR that predicts grass growth for dairy farmers. The authors aimed to provide users with high accuracy decisions and post-doc explainability by using the same district and farm data in the development of the system.

Kundu et al. \cite{kundu2021iot} developed an IoT-based automated data collection and classification framework for plant disease detection. Custom-Net, which is created from state-of-the-art models such as ResNet, Inception, and VGG, was used for disease detection. The authors used GradCAM for model explainability.
Viana et al. \cite{viana2021evaluation} proposed a framework using explanatory ML to effectively plan and manage the use of wheat, maize, and olive land. In this framework, the effects of features such as slope, soil type, and drainage density as well as socio-economic conditions were investigated. The authors used RF algorithm as an ML model and permutation feature importance (PFI), partial dependence plots (PDPs) and LIME for explainability.
Garrido et al. \cite{garrido2022evaporation} focused on developing multivariate and explainable ML models to predict evaporative water loss in irrigated agriculture. For this purpose, the ANN model, which takes climate variables such as soil, temperature, solar heating, and pressure as input, has been proposed. Model explainability was performed based on rule inference based on DT.

\begin{table*}[htb]
\centering
\caption{XAI Methodology Mapping by IoT Application Domain}
\label{tab:IoT_XAI_maping}
\resizebox{0.95\textwidth}{!}{%
\begin{tabular}{|l|l|l|l|l|l|l|l|l|}
\hline
\textbf{} & \textbf{\begin{tabular}[c]{@{}l@{}}Autonomous \\Systems and\\ Robotics\end{tabular}} & \textbf{\begin{tabular}[c]{@{}l@{}}Energy \\ Management\end{tabular}} & \textbf{Environment} & \textbf{Finance} & \textbf{Healthcare} & \textbf{Industrial} & \textbf{\begin{tabular}[c]{@{}l@{}}Security and \\ Privacy\end{tabular}} & \textbf{\begin{tabular}[c]{@{}l@{}}Smart \\ Agriculture\end{tabular}} \\ \hline
\textbf{Transparent/Rule Extraction} &  &  &  & \cite{sachan2020explainable} & \cite{hatwell2020ada}& \cite{mehdiyev2021explainable} & \cite{mahbooba2021explainable} & \cite{garrido2022evaporation} \\ \hline
\textbf{Example-based} &  &  &  &  &  &  & \cite{zolanvari2021trust} &  \\ \hline
\textbf{Feature Importance} & \cite{guo2020partially} & \cite{sirmacek2020occupancy,amiri2021peeking},  & \cite{kalamaras2019visual,ryo2021explainable,graham2020genome} &  \cite{bussmann2021explainable}, \cite{gramegna2020buy}, \cite{gite2021explainable}& \multicolumn{1}{c|}{\cite{pnevmatikakis2021risk,gozzi2022xai,dave2020explainable}} & \begin{tabular}[c]{@{}l@{}}\cite{rehse2019towards}, \cite{chen2020vibration,sun2020vision},\cite{senoner2021using}\\ \cite{serradilla2020interpreting,brito2022explainable}\end{tabular} & \cite{khan2021new,sarhan2021explainable,le2022classification} &\cite{kundu2021iot,viana2021evaluation} \\ \hline
\textbf{Gradient/Backpropagation} & \cite{iyer2018transparency} &  & \cite{diallo2020explainable} &  & \cite{gozzi2022xai} &  & \cite{saharkhizan2020ensemble,nascita} &  \\ \hline
\textbf{Perturbation} &  &  &  & \cite{gite2021explainable} & \cite{hossain2020explainable} &  & \cite{khan2021new} &  \\ \hline
\textbf{Visualization} &  & \cite{kim2019electric} &  &  & \cite{chittajallu2019xai},\cite{hossain2020explainable,monroe2021hiho}  &  &  &  \\ \hline
\end{tabular}%
}
\end{table*}

\section{Challenges, Open Issues, and Future Research Directions}
\label{Sec5_OpenC_FutureD}
There is ample evidence that humans have over-trusted AI systems in the past, and they still have a long way to go before they can fully trust them today. \cite{ghassemi2021false}. Apart from the benefits and potential that XAI brings to the realm of complex decision systems, it also has certain drawbacks \cite{watson2020conceptual, das2020opportunities, arrieta2020explainable} which are listed below:
\begin{itemize}

\item Within the scope of XAI, some key concepts are contradictory or imprecise. It has non-standardized terminology. Everyone agrees, for example, that algorithmic explanations should be faithful. However, it is unclear whether the loyalty should be to the target model or to the data generation process. These conceptual dilemmas cause misunderstanding and pointless debate \cite{watson2020conceptual}.
\item The bulk of XAI methods do not measure expected error rates. This makes it difficult to subject algorithmic explanations to severe tests, as required by any scientific hypothesis \cite{watson2020conceptual}.
\item The absence of a quantitative measure of the completeness and accuracy of interpretable systems. When assessing the quality of interpretability of a model, it should not vary according to the field knowledge of the observer. It is necessary to determine the most appropriate objective measurement metrics \cite{das2020opportunities}.
\item When ML results are more explicitly stated in terms of how they behave for a specific problem, malicious people can use this information for their own benefit. An attacker, for example, can understand what information an ML algorithm utilizes to arrive at a specific result via interpretability. With this knowledge, it attempts to manipulate the ML algorithm by seeking to find minor changes in the inputs that will result in a different output result \cite{arrieta2020explainable}. 
\item Some commonly used XAI models are vulnerable to adversarial attacks and this raises the concerns about should we trust the XAI models if it is manipulated \cite{fidel2020explainability}.
\item There is a lack of proper structure to combine multiple XAI methods with the aim of generating more complete explanations. 
\item Explanations can be illogical for the non-experts. An explanation may neglect too many features, which goes beyond the purpose of models. It explains what are important features, but does not explain what is the exact relationships between features and why they are important. The explanations are required for looking into the domain experts or some techniques to find out the correlations between the features and what they mean for the overall result.
\end{itemize}
In the light of the above challenging issues, some future directions can be listed as follows to be a guide for the related users.
\begin{itemize}
    \item For complete and clear explanations, model-specific methods should be used for improved fidelity since they are capable of looking into deeper the model architecture and specialized for an explanation of the inner structure about how it reaches the decision. Also, combining methods can overcome the partial explanations and provide a complete explanation for the model.
    \item For easier explanations, automation of XAI models can be provided as ML does recently, performing certain tasks such as feature selection, and hyperparameter tuning. In addition, advances in technology will allow for improvements in libraries and packages, which enables a high level of explanations.
    \item Trusting the results of XAI becomes a growing issue as the applications of IoT become wider. Therefore, improving security and resiliency of XAI models can be an emerging topic to deal with. The issues such as security, accountability, fairness and ethics are discussed in the context of Responsible AI by the authors of \cite{arrieta2020explainable}.
    \item IoT network usually poses a distributed architecture where IoT data is collected and processed by fog or cloud servers. They collaborate with AI systems. By exploiting federated learning, fog servers have potential to explain black-box models locally and generate local decisions, while cloud servers explain models globally and generate aggregated explanations. 
    \item IoT has various application areas such as wearable systems or nano-based IoT. XAI may use application-specific interfaces so that the model can understand the details of applications and generates granular explanations.
\end{itemize}

\section{Conclusion}
\label{Sec6_Conclusion}
In this paper, we present a detailed survey of XAI models in AI-powered IoT applications. Our study provides researchers with a comprehensive perspective on the usage areas of XAI methods in the IoT domain. In this context, we first explain the usage requirements and potential benefits of XAI methods in IoT. Then, we comprehensively examine XAI studies in IoT by application areas. Also, we summarize all the studies with a broad perspective in accordance with the XAI terminology and taxonomy. We finally present innovative ideas for potential future studies by focusing on future direction and open issues.

\ifCLASSOPTIONcaptionsoff
  \newpage
\fi

\bibliography{references.bib}

\bibliographystyle{IEEEtran}
\vskip -2\baselineskip plus -1fil

\begin{IEEEbiography}[{\includegraphics[width=1in,height=1.35in,clip,keepaspectratio]{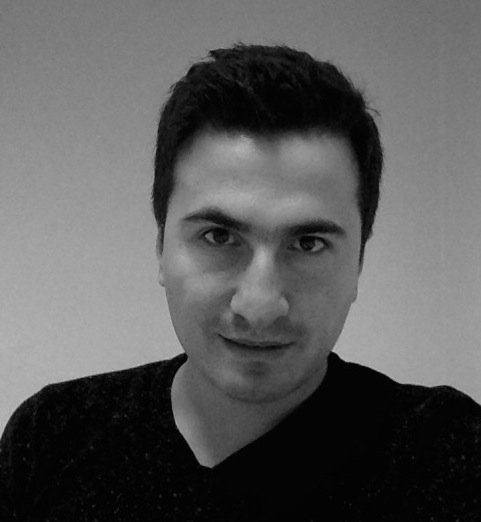}}]{İbrahim Kök}
	received his MSc and PhD degrees in Computer Science from Gazi University in 2015 and 2020, respectively. He is currently Assistant Professor with the Department of Computer Engineering, Pamukkale University, Denizli. His current research interests include Internet of Things (IoT), Deep Learning, AI-enabled IoT, and Data Analytics.
\end{IEEEbiography}
\vskip -2,5\baselineskip plus -1 fil

\begin{IEEEbiography}[{\includegraphics[width=1in,height=1.35in,clip,keepaspectratio]{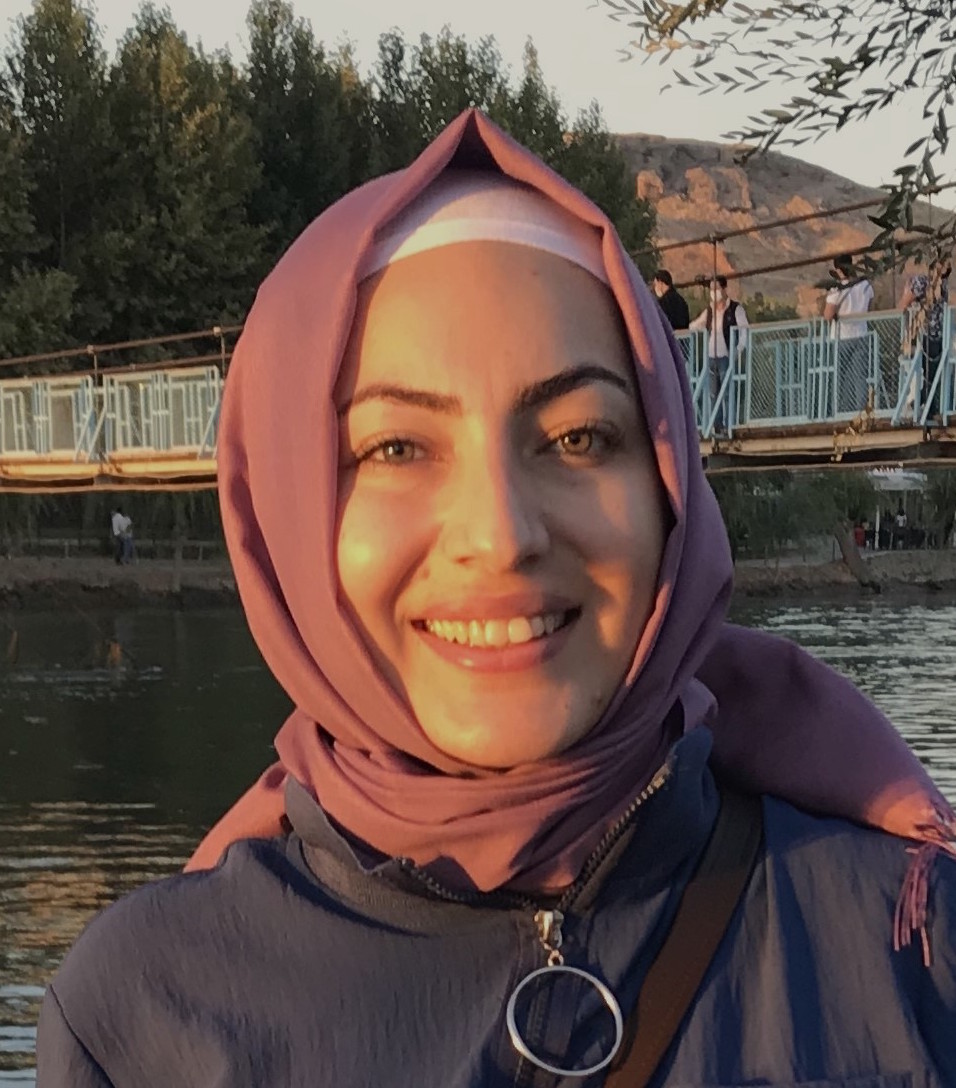}}]{Feyza Yıldırım Okay}
	received the MSc and PhD degrees in Computer Engineering from Gazi
    University Graduate School of Natural and Applied Sciences in 2013 and 2019, respectively. She is currently working as a research assistant at the same institute. Her research interests include Internet of Things, Fog Computing, Software Defined Networking, and Network Security.
\end{IEEEbiography}
\vskip -2,5\baselineskip plus -1 fil

\begin{IEEEbiography}[{\includegraphics[width=1in,height=1.35in,clip,keepaspectratio]{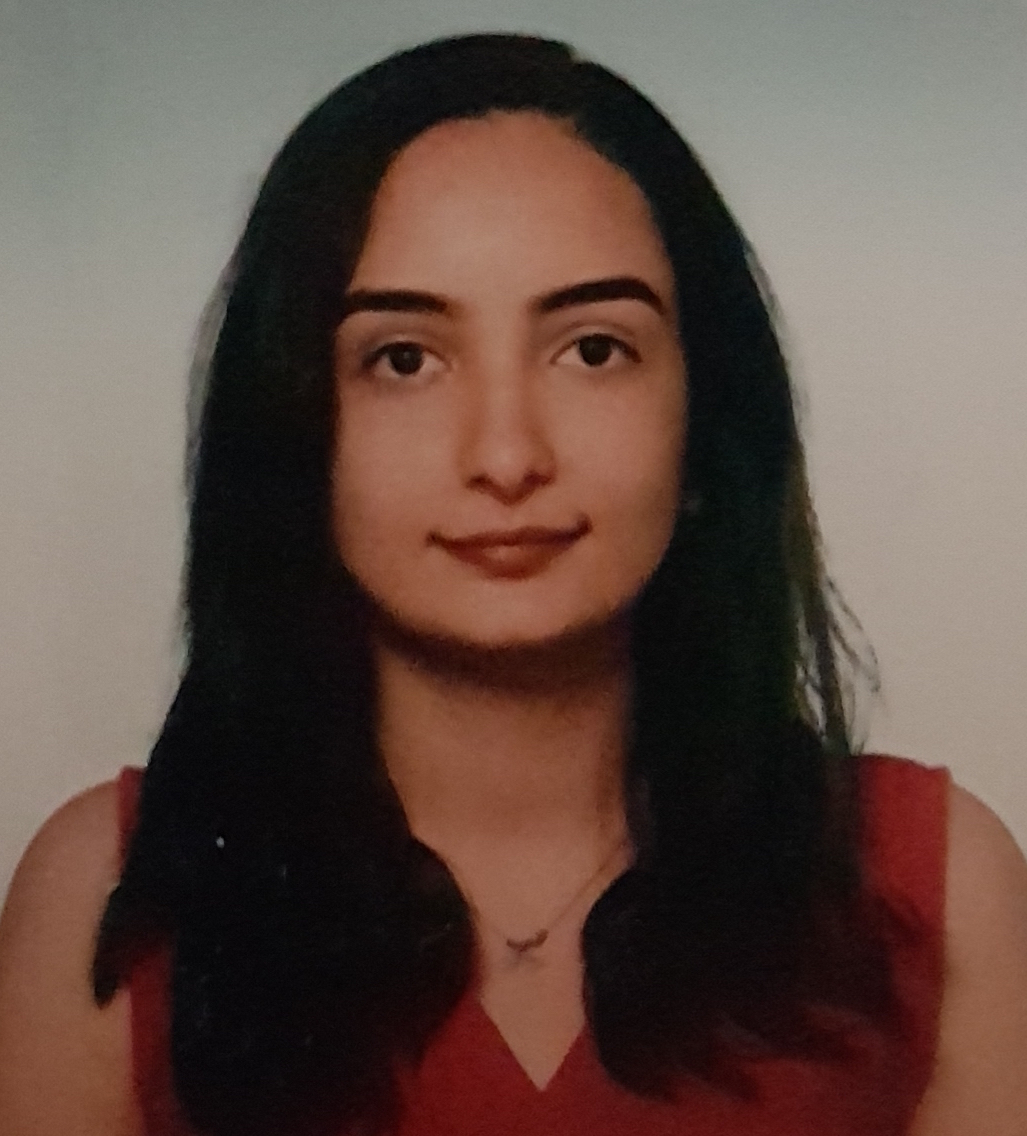}}]{Özgecan Muyanlı} received her BS in Computer Engineering from Gazi University in
2020. She is currently a PhD student with the Department of Computer Engineering, Hacettepe
University and working as a software engineer at TUSAŞ. Her research interests include Artificial
Intelligence, Data Analytics, Big Data, and Internet of Things.
	
\end{IEEEbiography}
\vskip -2,5\baselineskip plus -1 fil

\begin{IEEEbiography}[{\includegraphics[width=1in,height=1.35in,clip,keepaspectratio]{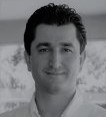}}]{Suat Özdemir}
    is with the Department of Computer Engineering at Hacettepe University, Ankara, Turkey. He received his MSc degree in Computer Science from Syracuse University (August 2001) and PhD degree in Computer Science from Arizona State University (December 2006). His current research interests include Internet of Things, Data Analytics, Artificial Intelligence, and Network Security.  \end{IEEEbiography}
  
\end{document}